\documentclass{article}
\usepackage{spconf,amsmath,amsfonts,graphicx,float}
\usepackage{tikz,tikzscale,pgfplots,filecontents}
\usepackage{multirow}
\usepackage{CJKutf8}
\usepackage{hyperref}
\usepackage{cite}

\title{Personalized word representations Carrying Personalized Semantics Learned from Social Network Posts}
\name{Zih-Wei Lin\sthanks{Fist author and second author are equal in contribution.}, Tzu-Wei Sung\footnotemark[1], Hung-Yi Lee, and Lin-Shan Lee}
\address{National Taiwan University\\
$\{$r04942111, b03902042, hungyilee$\}$@ntu.edu.tw, lslee@gate.sinica.edu.tw}

\begin{document}
%
\maketitle

\begin{abstract}
Distributed word representations have been shown to be very useful in various natural language processing (NLP) application tasks. These word vectors learned from huge corpora very often carry both semantic and syntactic information of words. However, it is well known that each individual user has his own language patterns because of different factors such as interested topics, friend groups, social activities, wording habits, etc., which may imply some kind of personalized semantics. With such personalized semantics, the same word may imply slightly differently for different users. For example, the word "Cappuccino" may imply "Leisure", "Joy", "Excellent" for a user enjoying coffee, by only a kind of drink for someone else. Such personalized semantics of course cannot be carried by the standard universal word vectors trained with huge corpora produced by many people. In this paper, we propose a framework to train different personalized word vectors for different users based on the very successful continuous skip-gram model using the social network data posted by many individual users. In this framework, universal background word vectors are first learned from the background corpora, and then adapted by the personalized corpus for each individual user to learn the personalized word vectors. We use two application tasks to evaluate the quality of the personalized word vectors obtained in this way, the user prediction task and the sentence completion task. These personalized word vectors were shown to carry some personalized semantics and offer improved performance on these two evaluation tasks.
\end{abstract}
\section{Introduction}
\label{sec:intro}
In many natural language processing tasks, a word is a discrete token and usually represented as a vector with one-hot encoding, where the dimensionality of the vector is the vocabulary size and the position of one corresponds to the index of the word in the vocabulary. One well-known limitation of such one-hot encoding method is that it says nothing regarding the semantic relationship between words. Various approaches to learn distributed word representations have been proposed to partly solve this problem~\cite{bengio2003neural, collobert2008unified, turian2010word, huang2012improving, mikolov2010recurrent, levy2014dependency, pennington2014glove}. Word2vec~\cite{mikolov2013efficient,mikolov2013distributed} is an unsupervised approach which has been shown to offer word representations carrying plenty of syntactic and semantic information, and found very useful in many applications such as identifying words with given semantics~\cite{mikolov2013linguistic, baroni2014don, levy2014neural}.

On the other hand, it is well known that each individual user has his own language patterns because of different factors such as interested topics, friend groups, social activities, wording habits, etc., which may imply some kind of personalized semantics. With such personalized semantics, the same word may imply slightly differently for different users. For example, the word "Cappuccino" may imply "Leisure", "Joy", "Excellent" for a user enjoying coffee, by only kind of drink for someone else. Such personalized semantics will certainly be helpful in improving the performance of the various natural language processing applications for each individual user. In fact personalization has been an important trend for many Internet services today, for example personalized retrieval~\cite{speretta2005personalized, fu2016enabling, shen2005implicit, xue2009user, chirita2007personalized}, personalized learning~\cite{su2015recursive, chen2010personalised}, and personalized recommendation systems~\cite{koren2009matrix, walter2008model, yang2014survey, deng2014social, park2007location, cho2002personalized}. An important step towards such personalized services is the personalized language processing~\cite{huang2014enriching, younus2014language, wen2012personalized, wen2013recurrent, tseng2015personalizing, lee2017personalizing, Halteren2004LinguisticPF, 1871}. However, the standard universal word representations learned from huge corpora produced by many people are certainly not able to describe personalized semantics. As a result, word representations adapted to different users is definitely a good step toward such a direction.

Substantial works have been reported on different ways for representing words as vectors to deal with different natural language processing problems~\cite{chen2014unified, luong2013better, santos2014learning, maas2011learning, cho2014learning, tang2014learning, ling2015finding, joulin2016bag}, but much less works were reported to investigate the mismatch between the universal word representations learned from general corpora and the personalized corpus produced by different individual users. One good reason for this is perhaps the difficulty in collecting personalized corpus. However, this situation has changed in recent years. Nowadays, many individuals post large quantities of texts over social networks, which can be a good source for constructing personalized corpus. In a series of efforts towards this direction, we implemented a cloud-based application to collect personalized linguistic data produced by many individual users from the social media. The data collected in this way are usually casual and short, but may carry plenty of personalized semantics.

In this paper, we proposed two approaches based on the skip-gram model of Word2vec to obtain personalized word vectors using individual social posts. The first approach simply tries to retrain the universal Word2vec model with the personalized corpus, while the second approach tries to insert an adaptive linear transformation layer within the skip-gram model. In both approaches, an universal Word2vec model was first trained with the background corpora produced by many people, then this Word2vec model was fine-tuned to be adapted to the personalized corpus. We used two different tasks to evaluate the quality of the obtained personalized word vectors, with which improved performance was obtained. We also found the second approach of inserting an adaptive linear transformation layer performed better.

\section{Proposed Approach}
\label{sec:approach}
We first illustrate the scenario of personalized word vectors in Subsection~\ref{ssec:scenario}, and briefly summarize the training criterion of skip-gram model in Subsection~\ref{ssec:skipgram}.  We then describe the two proposed approaches to train personalized word vectors for each individual user in Subsections~\ref{ssec:retrain} and~\ref{ssec:adaptive}.

\subsection{Scenario of Personalized Word Vectors}
\label{ssec:scenario}

\begin{figure}[htb]
  \centering
  \centerline{\includegraphics[width=0.9\linewidth]{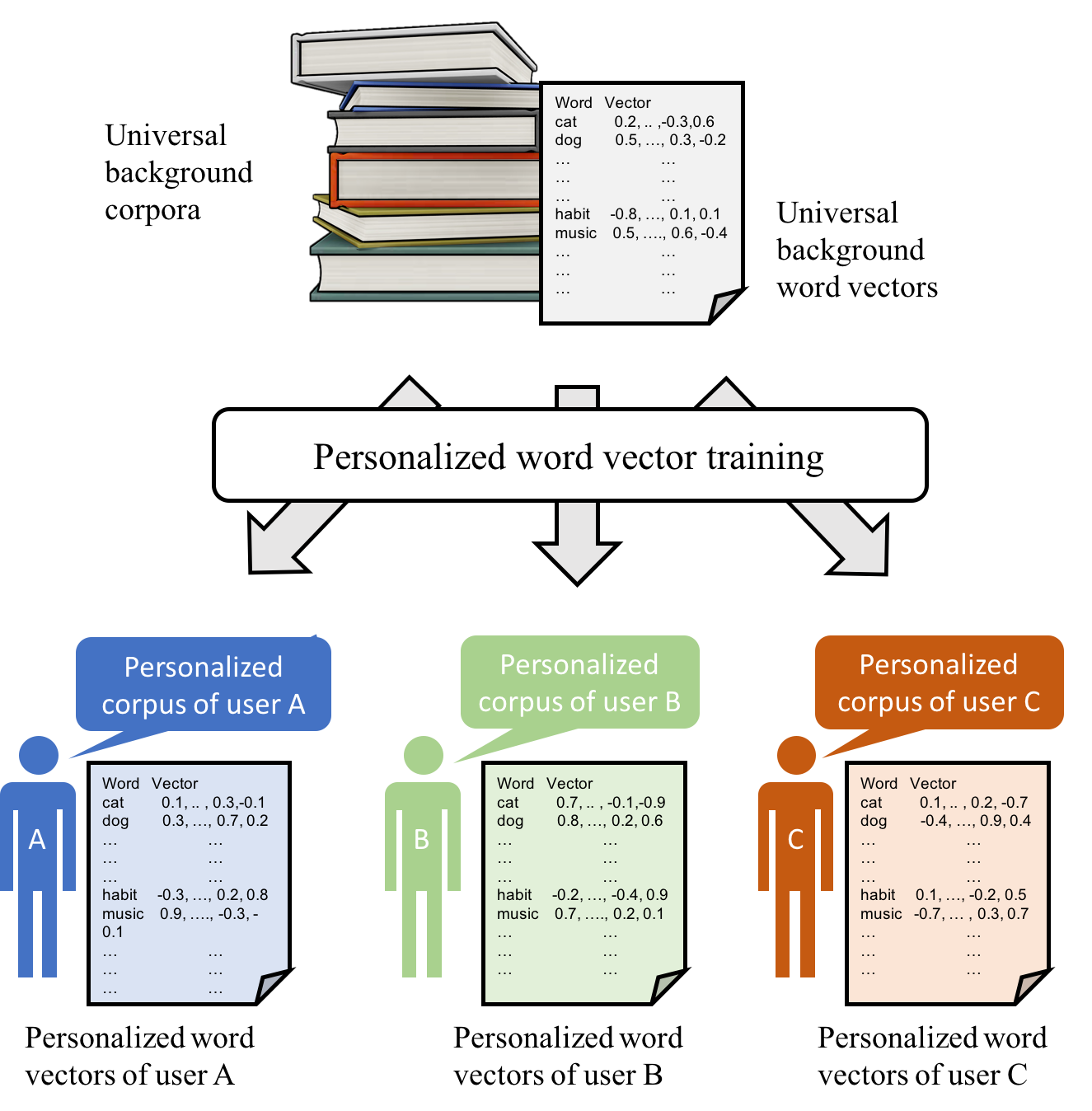}}
  \caption{Scenario of personalized word vectors.}
  \label{fig:Scenario}
\end{figure}

The scenario of personalized word vectors is shown in Fig.~\ref{fig:Scenario}. Universal background corpora including  numerous articles collected from different domains are first used to train a set of universal background word vectors using the skip-gram model. For each individual user, we then collect his (or her) social posts from the social media taken as the personalized corpus, with which we tune the universal background word vectors to obtain the personalized word vectors. The personalized word vectors are based on the same lexicon as used in the background corpora, but they are different in vector representations. These personalized word vectors are then used in various natural language processing applications.

\subsection{Skip-gram Model}
\label{ssec:skipgram}

\begin{figure*}[t]
	\centering
	\begin{minipage}[b]{.48\linewidth}
		\centering
  		\centerline{\includegraphics[width=0.9\linewidth]{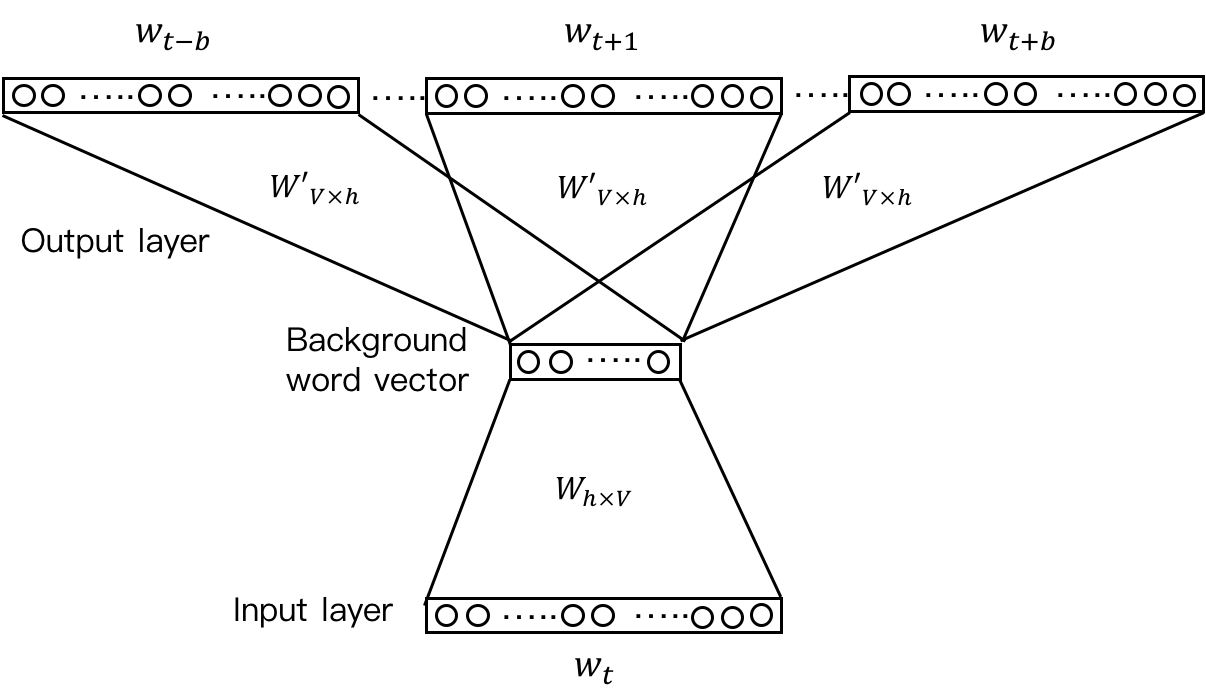}}
        \centerline{(a) Skip-gram model}\medskip
	\end{minipage}
	\begin{minipage}[b]{.48\linewidth}
		\centerline{\includegraphics[width=0.9\linewidth]{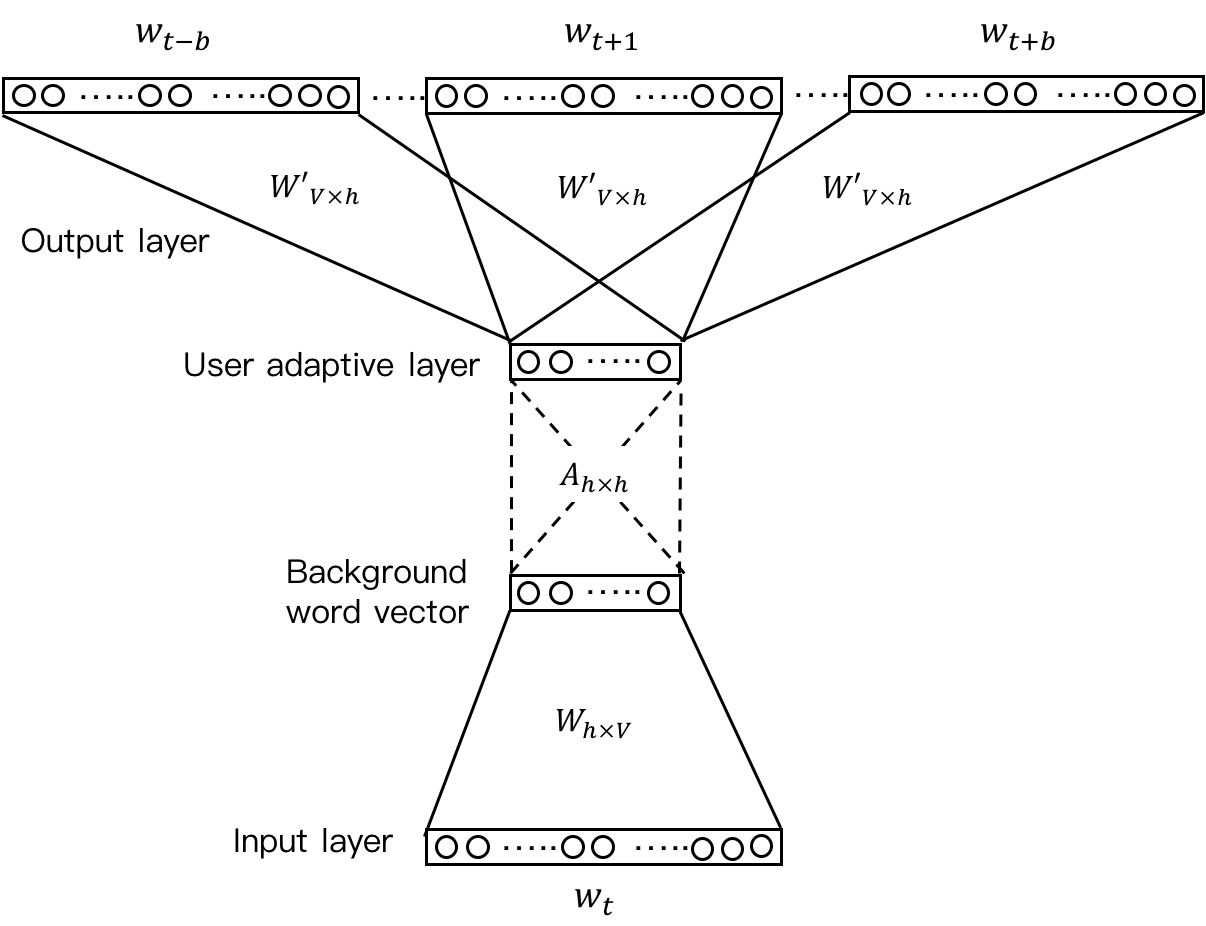}}
        \centerline{(b) Inserting a user adaptive layer to the skip-gram model}\medskip
	\end{minipage}
  \caption{Skip-gram model and inserting a user adaptive layer to the skip-gram model.}
  \label{fig:model-structure}
\end{figure*}

In this work, we choose the skip-gram model to train the word vectors. Given a sequence of training words $w_1, w_2, ..., w_T$, and the contexts $w_j$ for each word $w_t$, $t-b \le j \le t+b, b\neq 0$, where the context window length is $2b+1$, the goal of the skip-gram model as shown in Fig.~\ref{fig:model-structure} (a) is to find the parameters $\mathcal{W}, \mathcal{W}'$ so as to maximize the log of the conditional probability
\begin{equation}\label{eq:skipgram}
\underset{\mathcal{W}, \mathcal{W}'}{\arg\max} \sum_{t=1}^{T} \sum_{j=t-b,b\neq 0}^{t+b} \log{p}(w_{j}|w_{t};\mathcal{W}, \mathcal{W}')\quad.
\end{equation}

To approximate the conditional probability $p(w_{j}|w_{t};\mathcal{W}, \mathcal{W}')$ in Eq.(\ref{eq:skipgram}), negative sampling can be used to optimize the model parameters $\mathcal{W}, \mathcal{W}'$ so as to minimize the objective function for each word $w_t$, $J(w_t)$, defined as
\begin{equation}\label{eq:objective}
\resizebox{0.91\hsize}{!}{%
$J(w_t)=-\log{(\frac { 1 }{ 1+{ e }^{ -v'_{w_j}\cdot v_{w_t} } } )} - \sum\limits_{neg}{\log{(1-\frac { 1 }{ 1+{ e }^{ -v'_{w_{neg}} \cdot v_{w_t} } })} }$,%
}
\end{equation}
where $v_{w_t}$ and $v'_{w_j}$ are the vector representations for the target word $w_t$ and contexts $w_j$, and $v'_{w_j}$ is also called positive example. ${neg}$ is a function which randomly samples words $w_{neg}$ from the whole corpus, which are different from $w_j$ and called negative examples, according to their word frequencies. Empirically, $w_{neg}$ is picked from the distribution $U(w)^{\frac{3}{4}} / Z$, where $U(w)$ is the unigram distribution of the corpora, and $Z$ is a normalization constant. The goal of this objective function $J(w_t)$ in Eq.(\ref{eq:objective}) is to increase the quantity of $v'_{w_j}\cdot v_{w_t}$ for word-context pairs, and decrease $v'_{w_{neg}}\cdot v_{w_t}$ for randomly sampled irrelevant pairs. Therefore, vectors of words that share many contexts will be clustered together, and as a result these vectors can exhibit some semantics including the linear structure that makes analogical reasoning possible. 

\subsection{Approach 1 - Retrain the Model}
\label{ssec:retrain}
With the universal background word vectors trained with the skip-gram model as mentioned above using the universal background corpora, for each user, we retrain the background word vectors with the personalized corpus for each user with the same model, but simply fine-tune the parameters of the model to fit the personalized corpus. The fine-tuned word vectors are the personalized word vectors.

\subsection{Approach 2 - Inserting a User Adaptive Layer}
\label{ssec:adaptive}
This approach is shown in Fig.~\ref{fig:model-structure} (b), which is very similar to the skip-gram model in Fig.~\ref{fig:model-structure} (a), except we insert a user adaptive layer, which is a linear layer, between the original hidden and output layers.

As shown in Fig.~\ref{fig:model-structure} (b), we first train the background word vectors with the standard skip-gram model. This includes the input layer weights $\mathcal{W}_{h\times V}$ and output layer weights $\mathcal{W}'_{V\times h}$, where $h$ is the dimensionality of the word vectors, and $V$ is the vocabulary size. These weights are trained with the universal background corpora. Then the additional user adaptive layer is inserted into the model with weights $\mathcal{A}_{h\times h}$, where the weights $\mathcal{A}_{h\times h}$ are randomly initialized. We now fix the parameters for the universal background model, $\mathcal{W}_{h\times V}$ and $\mathcal{W}'_{V\times h}$, but only $\mathcal{A}_{h\times h}$, or the user adaptive matrix, is fine-tuned for each user based on the personalized corpus. The training algorithm is the same as that in Section~\ref{ssec:skipgram}. We wish to find the best parameters $\mathcal{A}$ to maximize the conditional probability
\begin{equation}
\underset{\mathcal{A}}{\arg\max} \sum_{t=1}^{T} \sum_{j=t-b,b\neq 0}^{t+b} \log{p}(w_{j}|w_{t};\mathcal{A}, \mathcal{W}, \mathcal{W}')
\end{equation}
for each individual user. We finally multiply the background word vectors $\mathcal{W}_{h\times V}$ by the adaptive weights $\mathcal{A}_{h\times h}$ to obtain the personalized word vectors for each individual user.

\section{Evaluation Tasks}
Given a set of word representations or embeddings $\{v_1, \dots, v_V\}$ for the corresponding vocabulary $\{w_1, \dots, w_V\}$, where the vector representation of $w_i$ is $v_i$, where $V$ is the vocabulary size, so $M=\{(w_1, w_2, \dots, w_V):(v_1, v_2, \dots, v_V) \}$ is a mapping or the word representation being considered. Our goal is to evaluate whether $M$ is a ``good'' representation, or a ``good'' set of embeddings. 

We introduce here two tasks to perform the evaluation.
\subsection{User Prediction}
\label{ssec:up}
Assume a user produces a document of $N$ sentences, $D=\{s_1, s_2, \dots, s_N\}$, where $s_n$ is the n-th sentence, $1 \le n \le N$. We wish to predict the user producing this document out of a group of users $U=\{u_1, u_2, \dots\}$, each user $u$ having a personalized word representation or mapping $M_u$.

\subsubsection{Document Classification Approach}
The approach proposed to perform document classification with respect to different domains using Word2vec~\cite{DBLP:journals/corr/Taddy15} by maximizing the log likelihoods of words and their contexts can be used here, except each domain corresponds to a user. This is parallel to the objective function defined in Subsections ~\ref{ssec:skipgram} and~\ref{ssec:adaptive}. 

Consider a sentence $s_n$ with $L$ words, $s_n=[w_{n_1}, \dots, w_{n_L}]$, the log likelihood of $s_n$ based on the mapping or word representation $M=\{(w_1, \dots, w_V):(v_1, \dots, v_V) \}$ is
\begin{equation}
	\log p_M(s_n) = \sum_{t=1}^L\sum_{j=t-b,b\neq 0}^{t+b} \log {p_M(w_{n_{j}} | w_{n_t})}\quad , 
\end{equation}
where $w_{n_t}$ is the $n_t$-th word and $w_{n_j}$ is its context word, and $p_M(w_{j} | w_t)$ can be obtained with the mapping $M$,
\begin{equation}
	p_M(w_{j} | w_t) = \frac{e^{v'_j \cdot v_t}}{\sum_{i=1}^V{e^{v'_i \cdot v_t}}}\quad ,
\end{equation}
where $v_t$ is the representation of $w_t$ and so on, and the summation in the denominator is over all words in the vocabulary. So the document $D$, $D=\{s_1, \dots, s_N\}$, has log likelihood
\begin{equation}
	\log p_M(D) = \sum_{i=1}^{N}\log p_M(s_i)\quad .
\end{equation}
The posterior probability $p(u|D)$ that $D$ is produced by user $u$ can be derived from Bayes rule as follows:
\begin{equation}\label{eq:posterior}
	p(u|D) = \frac{p_{M_u}(D) \pi_u}{\sum_{u' \in U}p_{M_{u'}}(D) \pi_{u'}} 
\end{equation}
where $\pi_u$ is the prior probability of user $u$, $M_u$ is the personalized word vectors for user $u$, and the summation in the denominator is over all users considered.

Finally, the predicted user is $\hat{u}$:
\begin{equation}
	\hat{u} = \underset{u}{\arg\max} \: {p(u|D)}\quad.
\end{equation}

\subsubsection{Evaluation measure}
Two measures are used here:
\begin{enumerate}
\item Prediction accuracy: Percentage of documents for which the corresponding user is correctly predicted.
\item Mean reciprocal rank (MRR): If the correct user is predicted as the $r$-th candidate, the reciprocal rank is $\frac{1}{r}$. The mean reciprocal rank is the average of the reciprocal ranks so MRR should be less than $1.0$, and the closer to $1.0$ the better. 
\end{enumerate}

\subsection{Sentence Completion} 
\label{ssec:sc}
In this task, from each test sentence we scoop the word with maximum \texttt{TF-IDF}, and then use the semantics from the word vectors to find the best word to fill up the blank. This can be achieved by taking the average of embeddings of the remaining words in the sentence, then ranking all words based on the cosine similarity with respect to this average. The scooped word is taken as the correct answer and the mean reciprocal rank (MRR) is used in the evaluation. Higher MRR implies the word embedding is better.

\section{Experimental Setup}
\label{sec:}
\subsection{Corpus} \label{ssec:corpus}
First of all, 2.6M sentences including 42,558 distinct words in lexicon were collected from Plurk, a popular social networking site. These data from Plurk were used as the universal background corpora for training the universal background word vectors. The testing experiments were conducted on a set of personalized corpus crawled from Facebook. In order to obtain the personalize Facebook posts, we implemented a cloud-based application capable of helping users to access their social network via voice. Each user can log in his Facebook account and grant our application the authority to collect his linguistic data for experiment purposes. A total of 40 users did so. As a result, all data accessible to the accounts of these 40 target users were collected. This resulted in a total of 67,656 sentences. The number of sentences for each user ranged from 308 to 5,140, with 10.6 words (Chinese or English or mixed) per sentence in average. For each target user, 3/5 of his corpus is taken as the training set, 1/5 as the validation set, and the rest 1/5 for testing.

The code-mixing phenomenon appears in the sentences collected from both Plurk and Facebook. Most sentences were produced in Chinese, but some words or phrases were naturally produced in English and embedded in the Chinese sentences. The mix ratio for the Chinese characters to English words in the Facebook data is roughly 10.5:1.

\subsection{User Prediction \& Sentence Completion}
In the user prediction task, we divide each target user's testing set into smaller documents, each containing at most 30 sentences, and the total number of documents is 473. Each testing document is labeled with the user who produced the document. The user prior probabilities $\pi_u$ in Eq.(\ref{eq:posterior}) is taken as uniform for all users. 

In the sentence completion task, we preprocessed all users' testing set by means of scooping words as mentioned in Subsection~\ref{ssec:sc}. In total, there are 13,512 sentences for testing.



\section{Experimental Results}
\label{sec:}

\subsection{User Prediction}
This is for the tests mentioned in Subsection~\ref{ssec:up}. Table~\ref{table:inversion-res-overall} shows the MRR and prediction accuracy averaged over the testing set for the two approaches discussed in Subsections~\ref{ssec:retrain}, \ref{ssec:adaptive} compared to a baseline. The first section (A) (No Background) is for the results when all personalized word vectors were trained directly with the personalized corpus only, without using the background corpora. The second section (B) (Retrain) and third section (C) (Adaptive Layer) are respectively for the two proposed approaches summarized in Subsections~\ref{ssec:retrain} and~\ref{ssec:adaptive}, all with word vector dimensionality of 128, 192 and 256.

We see from section (A) the word vectors trained with personalized corpus only without background corpora got the worst MRR and accuracy in the table, obviously because the personalized corpus is too sparse to offer reasonably good word vectors. With the help of the background corpora and the universal background word vectors, we see the MRR and accuracy were significantly better and increased as the embedding size went bigger in sections (B)(C). Comparing the two proposed methods, we see Adaptive Layer (section (C)) was clearly better than Retrain (section (B)) with the same embedding size. There can be at least two reasons for this. First, there are much more parameters to be trained for the Retrain approach, i.e., there are $V \times h \times 2$ parameters to be trained for the matrices $\mathcal{W}_{h\times V}$ and $\mathcal{W}'_{V\times h}$, where $V$ is vocabulary size and $h$ is the embedding size. In contrast, in Adaptive Layer approach only $h \times h$ parameters for the matrix $\mathcal{A}_{h\times h}$ are to be trained. The former is much larger because the vocabulary size $V$ is usually at the order of ten thousands and the embedding size is at the order of hundreds. So much more personalized data are needed for the Retrain approach to learn high quality personalized word vectors. Second, in Retrain approach, the vectors for those words appearing in the personalized corpus were fine-tuned to fit the personalized corpus. However, for those words not appearing in the personalized corpus, the corresponding word vectors were almost never trained and simply remained primarily unchanged from those learned from the universal background corpora. So the words were in fact divided into two separate groups, the unseen words trained with the background corpora and the observed words trained with the personalized corpus. In contrast, the Adaptive Layer approach used an additional linear transformation layer to adapt the whole set of word vectors according to the personalized corpus. In other words, the linear adaptive layer learned a full transformation matrix $\mathcal{A}_{h\times h}$, although small, which mapped the whole set of background word vectors to a new space of personalized semantics. This linear transformation also prevented the word vectors from overfitting to the personalized corpus.

\begin{table}[t]
\centering
\resizebox{\columnwidth}{!}{%
\begin{tabular}{|l|c|r|r|}
\hline
Approaches                          & \begin{tabular}[c]{@{}c@{}}Embedding \\ Size\end{tabular} & MRR & \multicolumn{1}{c|}{\begin{tabular}[c]{@{}c@{}}Prediction\\ Accuracy\end{tabular}} \\ \hline
\multirow{3}{*}{(A) No Background}& 128                        & 0.256                    & 0.140               \\ \cline{2-4} 
                          & 192                                 & 0.296                    & 0.204               \\ \cline{2-4}
                          & 256                                 & 0.336                    & 0.226               \\ \hline
\multirow{3}{*}{(B) Retrain}  & 128                             & 0.402                    & 0.309               \\ \cline{2-4} 
                          & 192                                 & 0.424                    & 0.340               \\ \cline{2-4} 
                          & 256                                 & 0.430                    & 0.340               \\ \hline
\multirow{3}{*}{(C) Adaptive Layer} & 128                       & 0.580                    & 0.485               \\ \cline{2-4} 
                          & 192                                 & 0.610                    & 0.523                \\ \cline{2-4} 
                          & 256                                 & 0.630                    & 0.512                \\ \hline
\end{tabular}%
}
\caption{Evaluation results for the user prediction task using different approaches, all with the personalized data.}
\label{table:inversion-res-overall}
\end{table}

\begin{figure}[h]
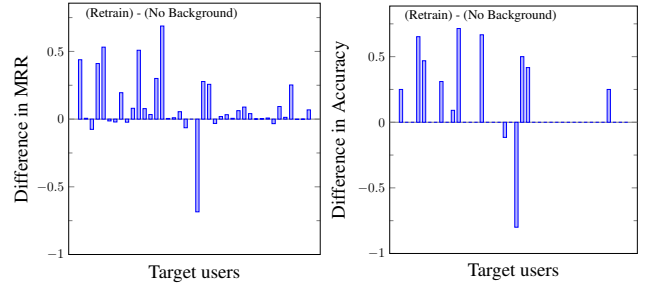

	\centering
	\begin{minipage}[b]{.48\linewidth}
      	\centering
      	\centerline{\includegraphics[width=\linewidth]{diff-mrr-2.tikz}}
        \centerline{(a) MRR Differences}\medskip
	\end{minipage}
	\begin{minipage}[b]{.48\linewidth}
		\centering
    	\centerline{\includegraphics[width=\linewidth]{diff-acc-2.tikz}}
        \centerline{(b) Prediction Accuracy Differences}\medskip
	\end{minipage}
	\caption{For the user prediction task: difference in (a) MRR and (b) Prediction Accuracy for each individual user for those obtained with Retrain approach minus No Background approach, all at embedding size of 256.}
    \label{fig:inversion-res-diff-retrain-wo-background}
\end{figure}

\begin{figure}[h]
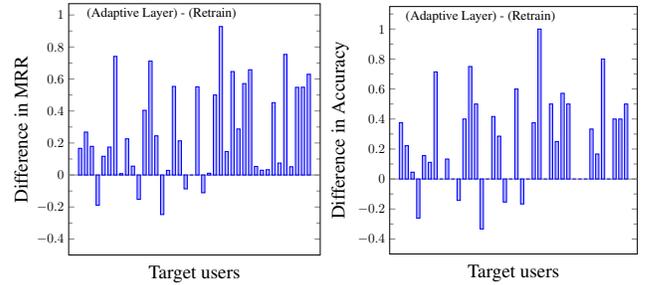

	\centering
	\begin{minipage}[b]{.48\linewidth}
      	\centering
      	\centerline{\includegraphics[width=\linewidth]{diff-mrr.tikz}}
        \centerline{(a) MRR Differences}\medskip
	\end{minipage}
	\begin{minipage}[b]{.48\linewidth}
		\centering
    	\centerline{\includegraphics[width=\linewidth]{diff-acc.tikz}}
        \centerline{(b) Prediction Accuracy Differences}\medskip
	\end{minipage}
	\caption{For the user prediction task: difference in (a) MRR and (b) Prediction Accuracy for each individual user for those obtained with Adaptive Layer approach minus Retrain approach, all at embedding size of 256.}
    \label{fig:inversion-res-diff}
\end{figure}

Since the averages didn’t actually tell how the different approaches compared with each other for each individual user, we plot in addition the differences in MRR and prediction accuracy across all the 40 target users in Figs.~\ref{fig:inversion-res-diff-retrain-wo-background} and~\ref{fig:inversion-res-diff}. Each bar in the figures represents the score obtained with one approach minus that with another, all with embedding size of 256. Fig.~\ref{fig:inversion-res-diff-retrain-wo-background} is for the Retrain approach minus No Background, while Fig.~\ref{fig:inversion-res-diff} is for the Adaptive Layer approach minus the Retrain approach. From Figs.~\ref{fig:inversion-res-diff-retrain-wo-background} and~\ref{fig:inversion-res-diff}, we see the differences are quite apparent for most target users. 

\subsection{Sentence Completion}
\label{ssec:}

\begin{table}[h]
\centering
\resizebox{\columnwidth}{!}{%
\begin{tabular}{|l|c|c|r|}
\hline
Approaches                          & \begin{tabular}[c]{@{}c@{}}Embedding\\ Size\end{tabular} & \begin{tabular}[c]{@{}c@{}}(1)\\ Percentage within\\ Top 500 (\%)\end{tabular} & \multicolumn{1}{c|}{\begin{tabular}[c]{@{}c@{}}(2)\\ MRR \\ within (1)\end{tabular}} \\ \hline\multirow{3}{*}{(A) No Background}  & 128            & 4.49         & 0.178                    \\ \cline{2-4} 
                                    & 192            & 4.50         & 0.168                    \\ \cline{2-4} 
                                    & 256            & 4.57         & 0.186                    \\ \hline
\multirow{3}{*}{(B) Background}     & 128            & 4.70         & 0.182                    \\ \cline{2-4} 
                                    & 192            & 4.73         & 0.194                    \\ \cline{2-4} 
                                    & 256            & 4.72         & 0.188                    \\ \hline
\multirow{3}{*}{(C) Retrain}        & 128            & 4.76         & 0.186                    \\ \cline{2-4} 
                                    & 192            & 4.77         & 0.196                    \\ \cline{2-4} 
                                    & 256            & 4.85         & 0.201                    \\ \hline
\multirow{3}{*}{(D) Adaptive Layer} & 128            & 4.93         & 0.198                    \\ \cline{2-4} 
                                    & 192            & 4.98         & 0.210                    \\ \cline{2-4} 
                                    & 256            & 5.18         & 0.224                    \\ \hline
\end{tabular}%
}
\caption{Results for the sentence completion task: (1) percentage of test sentences with correct answer within top 500 and (2) MRR averaged for those sentences in (1).}
\label{table:sf-res}
\end{table}

Table~\ref{table:sf-res} reports the MRR for four different approaches. The sections (A)(C)(D) are for exactly the same cases as those in Table~\ref{table:inversion-res-overall}, respectively for using personalized corpus alone, and the personalized word vectors by the proposed two approaches. The extra section (B) (Background) is for the word vectors trained with the background corpora only. Column (1) (Percentage within Top 500) lists the percentages of the test sentences for which the correct words for the blanks were ranked within the top 500 words found by the word vectors, and column (2) (MRR within (1)) reports the MRR values averaged over those sentences with the correct words ranked within 500. It can be found that the two proposed approaches (sections (C)(D)) performed significantly better with trends consistent with those observed in Table~\ref{table:inversion-res-overall}. Also, because many of the test sentences are very short with only a few words, so the sentence completion task is actually a very difficult tasks here. As a result, only 4.49\% - 5.18\% of them had correct words within 500, and the MRR obtained was not high.

\begin{figure*}[htb]
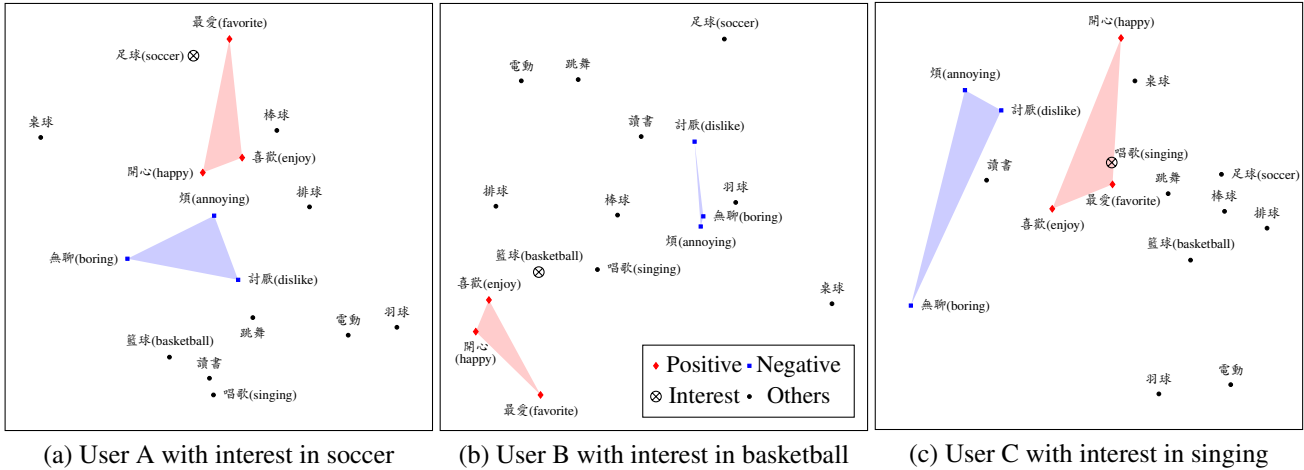

	\centering
	\begin{minipage}[b]{.32\linewidth}
      	\centering
      	\centerline{\includegraphics[width=\linewidth]{soccer-3.tikz}}
        \centerline{(a) User A with interest in soccer}\medskip
	\end{minipage}
	\begin{minipage}[b]{.32\linewidth}
		\centering
    	\centerline{\includegraphics[width=\linewidth]{basketball.tikz}}
        \centerline{(b) User B with interest in basketball}\medskip
	\end{minipage}
	\begin{minipage}[b]{.32\linewidth}
		\centering
    	\centerline{\includegraphics[width=\linewidth]{sing.tikz}}
        \centerline{(c) User C with interest in singing}\medskip
	\end{minipage}
   	\caption{Personalized word vectors visualization of three different users.}
    \label{fig:visualization}
\end{figure*}

\subsection{An Example}
We tried to visualize the personalized word vectors for three example users trained with the second approach of adaptive layer with dimensionality of 256, and plot small subsets of them with t-sne in Fig.~\ref{fig:visualization} (a)(b)(c) respectively. The black points marked by ``$\bullet$'' are Chinese words representing human activities such as singing(\begin{CJK}{UTF8}{bkai}唱歌\end{CJK}), dancing(\begin{CJK}{UTF8}{bkai}跳舞\end{CJK}), studying(\begin{CJK}{UTF8}{bkai}讀書\end{CJK}), basketball(\begin{CJK}{UTF8}{bkai}籃球\end{CJK}) and soccer(\begin{CJK}{UTF8}{bkai}足球\end{CJK}). The red triangle is a positive emotion triangle defined by three red points marked by ``$\Diamond$'' for words indicating positive emotion: happy(\begin{CJK}{UTF8}{bkai}開心\end{CJK}), favorite(\begin{CJK}{UTF8}{bkai}最愛\end{CJK}), enjoy(\begin{CJK}{UTF8}{bkai}喜歡\end{CJK}), while the blue triangle is a negative emotion triangle defined by three blue points marked by ``$\Box$'' for words indicating negative emotion: dislike(\begin{CJK}{UTF8}{bkai}討厭\end{CJK}), boring(\begin{CJK}{UTF8}{bkai}無聊\end{CJK}), annoying(\begin{CJK}{UTF8}{bkai}煩\end{CJK}). Fig.~\ref{fig:visualization} (a)(b)(c) are the word vectors for three different users A, B, C with different personal interests respectively in soccer(\begin{CJK}{UTF8}{bkai}足球\end{CJK}), basketball(\begin{CJK}{UTF8}{bkai}籃球\end{CJK}) and singing(\begin{CJK}{UTF8}{bkai}唱歌\end{CJK}), word vectors for which are respectively marked by  ``$\otimes$'' in the subfigures (a)(b)(c). In Fig.~\ref{fig:visualization} (a), user A is interested in soccer(\begin{CJK}{UTF8}{bkai}足球\end{CJK}). We can see his word vector for soccer(\begin{CJK}{UTF8}{bkai}足球\end{CJK}) is close to the positive triangle but far from the negative triangle. However, in Fig.~\ref{fig:visualization} (b)(c) for users B and C with different interests, the word soccer(\begin{CJK}{UTF8}{bkai}足球\end{CJK}) is more or less neutral in emotion. Similarly user B in Fig.~\ref{fig:visualization} (b) is interested in basketball(\begin{CJK}{UTF8}{bkai}籃球\end{CJK}), so the word basketball(\begin{CJK}{UTF8}{bkai}籃球\end{CJK}) in Fig.~\ref{fig:visualization} (b) is close to the positive emotion triangle but far from negative triangle, but is more or less neutral in Fig.~\ref{fig:visualization} (a)(c). Same in Fig.~\ref{fig:visualization} (c) for user C who is interested in singing(\begin{CJK}{UTF8}{bkai}唱歌\end{CJK}). These results demonstrate the approach proposed here is able to actually extract some personalized semantics as discussed earlier in this paper.

\section{Conclusions}\label{sec:}
In this paper, we proposed a new framework for training personalized word vectors carrying personalized semantics using personalized data crawled from social networks. The word vectors are first trained with universal background corpora to learn the general knowledge, and then adapted towards the personalized data by fine-tuning the background word vectors. Two approaches were proposed for the adaptation, one by retraining the word vectors while the other by inserting an adaptation layer. Experimental results over a user prediction task and a sentence completion task showed that both approaches offered consistently better results, and the adaptive layer approach is better than the retrain approach.


\bibliographystyle{IEEEbib}
\bibliography{strings,refs}

\end{document}